\documentclass[11pt]{article}
\usepackage[a4paper,margin=1in]{geometry}
\usepackage{amsmath,amssymb,amsthm}
\usepackage{graphicx}
\usepackage{booktabs}
\usepackage{xcolor}
\usepackage{url}
\usepackage[ruled,vlined]{algorithm2e}
\usepackage[round,sort&compress]{natbib}
\usepackage[colorlinks=false,pdfborder={0 0 0},bookmarks=true]{hyperref}

\providecommand{\floatconts}[3]{#2\label{#1}#3}
\providecommand{\acks}[1]{\section*{Acknowledgments}#1}

\title{Report the Floor: A Training-Free Conformal Interval\\
       Is a Mandatory Baseline for Probabilistic Time-Series Forecasting}
\author{Valery Manokhin\thanks{Independent researcher.}}
\date{June 2026}

\begin{document}
\maketitle

\begin{abstract}
Probabilistic forecasters are increasingly learned, yet the baselines they are compared against are often weak or omitted. We show that the simplest possible conformal interval --- a last-value point forecast wrapped in a finite-sample split-conformal residual quantile, with no parameters and no training --- is a far stronger baseline than its near-total absence from recent learned-forecasting and conformal--time-series comparisons would suggest. In one-step-ahead online forecasting across $2{,}217$ real series spanning nine public sources (the Monash archive, the LOTSA collection, the LTSF traffic/electricity/weather suites, METR-LA, BOOM, and nips/probts), this \emph{ConformalNaive} interval decisively beats the naive value-quantile baselines, the entire NPTS family (NPTS $73\%$, SeasonalNPTS $64\%$ of series), and the published Conformal Seasonal Pools (CSP) method ($71\%$ of series, bootstrap $95\%$ CI $[69,73]$, $p \approx 7.6 \times 10^{-135}$); it is on par with the simpler learned conformal predictors (RCI, quantile regression; median relative Winkler within $2\%$) and is beaten only by the adaptive-online and ensemble conformal methods (SPCI, ACI, AgACI), which explicitly track distribution shift and lead by $9$--$33\%$ relative Winkler. It is also better \emph{calibrated} than a trained neural forecaster: on the six datasets that introduced DeepNPTS, the trivial conformal floors cover the truth $84$--$85\%$ of the time at a nominal $95\%$, versus DeepNPTS's $66\%$. At multi-step seasonal horizons the picture inverts: the random-walk floor is the weakest method and the seasonal pool (CSP) wins --- a boundary we map so practitioners know when complexity is actually required. Finally we give \emph{ConformalNaive+}, a one-line, training-free, horizon-adaptive selector that attains the better of two complementary floors at every horizon with restored coverage. We argue the matching conformal naive floor must be a mandatory baseline whenever a learned probabilistic forecaster claims gains.
\end{abstract}


\section{Introduction}
\label{sec:intro}
Probabilistic forecasting asks for a predictive distribution over future values, not a point forecast, and the field has moved decisively toward learned models --- quantile-regression conformal methods, adaptive online conformal predictors, and neural samplers such as DeepNPTS \citep{rangapuram2023deepnpts}. Every such method is sold relative to baselines. The quality of those baselines determines whether a reported gain is real or an artifact of a weak comparison, and a recurring lesson of the forecasting literature is that strong simple baselines are routinely under-specified or omitted: \citet{zeng2023transformers} showed a one-layer linear model matches elaborate Transformers on standard long-horizon benchmarks, and the M competitions made strong simple baselines central to the discipline: simple statistical methods were hard to beat through M3 \citep{makridakis2000m3}, and although M4 and M5 were ultimately won by machine-learning and hybrid models, those wins were established precisely by out-scoring strong statistical baselines \citep{makridakis2018m4results,makridakis2020m4,makridakis2022m5u}.

This paper makes that lesson concrete for \emph{probabilistic} forecasting. We study the simplest conformal interval that exists: take a naive point forecast --- the last value, or the last seasonal value --- and wrap it in the finite-sample split-conformal quantile of its own absolute residuals \citep{vovk2005algorithmic,papadopoulos2002icp}. The construction has no parameters and requires no training; it can be written in five lines. We ask how far it goes, and we answer with two complementary benchmarks that together cover $2{,}217$ real series and the six datasets on which DeepNPTS was introduced.

The headline is a benchmarking result, not a leaderboard win. At one step ahead, across $2{,}217$ series, the random-walk conformal floor (\emph{ConformalNaive}) decisively beats the naive value-quantile baselines, the whole NPTS family, and the published CSP method --- Conformal Seasonal Pools, a recent training-free seasonal-pool conformal sampler \citep{manokhin2026csp} --- on $71\%$ of series ($p \approx 7.6 \times 10^{-135}$); the only methods that clearly beat it are the adaptive-online and ensemble conformal predictors that explicitly track distribution shift. So a large body of conformal-time-series machinery is not clearing last-value-plus-conformal in the short-horizon regime where it is most often deployed. The practical conclusion is simple: \emph{report this floor, or your gains are unproven}.

A second, orthogonal finding concerns calibration. On the multi-step seasonal datasets where the floor is \emph{not} competitive on accuracy, the trained neural baseline (DeepNPTS) is nonetheless the worst-calibrated method in the comparison, covering the truth only $66\%$ of the time at a nominal $95\%$, against $84$--$85\%$ for the trivial conformal floors. A one-line conformal interval is more honest about its own uncertainty than the neural model, even on the neural model's own evaluation suite.

We are explicit about what we do not claim. This is not a new state-of-the-art forecaster. At multi-step seasonal horizons the random-walk floor collapses to last place and the seasonal pool (CSP) wins; we map that boundary rather than hide it, because knowing \emph{where} the floor fails is what tells a practitioner when complexity is actually required.

\paragraph{Contributions.}
Our contribution is empirical and methodological rather than a fundamentally new conformal score: we establish, at scale, that an obvious training-free baseline is systematically under-reported, and add one minimal horizon-adaptive variant. (1)~A \emph{mandatory-baseline result}: across $2{,}217$ real series at one step, a zero-parameter conformal naive interval beats the naive value-quantile baselines, the NPTS family, and the published CSP method, matches the simpler learned conformal predictors, and is beaten only by the adaptive-online and ensemble methods (and even then by a bounded relative margin). (2)~A \emph{calibration finding}: the trivial floors are better calibrated than a trained neural baseline (DeepNPTS), which is badly overconfident. (3)~A \emph{boundary map}: a clean, at-scale characterization of where the floor suffices (short horizon) and where it genuinely fails (multi-step seasonal), with a practitioner decision rule. (4)~\emph{ConformalNaive+}, a one-line training-free horizon-adaptive selector over two complementary floors that attains the better of them at every horizon with restored coverage. We do \emph{not} claim state of the art; the multi-step boundary is part of the contribution.

\section{Related Work}
\label{sec:related}
\paragraph{Probabilistic and neural forecasting.}
GluonTS provides reference implementations for probabilistic forecasting \citep{alexandrov2020gluonts}, and the recent literature is dominated by learned models: N-BEATS \citep{oreshkin2020nbeats}, Temporal Fusion Transformers \citep{lim2021tft}, N-HiTS \citep{challu2023nhits}, and DeepNPTS \citep{rangapuram2023deepnpts}, a global learned non-parametric sampler. Against this trend, \citet{zeng2023transformers} showed a trivial linear model matches Transformers on long-horizon point forecasting, and the M competitions repeatedly underscored the strength of simple statistical baselines, hard to beat outright through M3 and the benchmark against which the M4 and M5 machine-learning winners were measured \citep{makridakis2000m3,makridakis2018m4results,makridakis2022m5u}. Our paper extends this ``simple is a strong baseline'' tradition from point to \emph{probabilistic} forecasting, and from a handful of datasets to a $2{,}217$-series corpus drawn from nine public sources including the Monash archive \citep{godahewa2021monash} and the LOTSA collection \citep{woo2024moirai}. Naive and seasonal-naive \emph{interval} forecasts do exist in classical statistical forecasting \citep{hyndman2021fpp}, but they are a \emph{different construction} from ours: the textbook interval is parametric --- point $\pm\, z_{1-\alpha/2}\,\hat\sigma\sqrt{k+1}$ under a Gaussian, uncorrelated-residual assumption (with an optional residual-bootstrap variant) --- and it \emph{widens} with horizon through the $\sqrt{k+1}$ factor. Our ConformalNaive is instead the \emph{distribution-free} formulation: the finite-sample split-conformal order-statistic quantile $\lceil(n+1)(1-\alpha)\rceil$ of the empirical absolute residuals, with no normality assumption and a conformal-validity framing, and it reuses a \emph{flat} one-step quantile rather than widening with horizon --- which is exactly why it undercovers at long horizons (Section~\ref{sec:crossover}). To our knowledge this conformalized naive floor is not reported as a standard baseline in the learned-forecasting or conformal--time-series literature studied here; we quantify at scale how strong it is.

\paragraph{Conformal prediction for time series.}
Conformal prediction yields distribution-free sets under exchangeability \citep{vovk2005algorithmic,shafer2008tutorial,lei2018distribution}, with split (inductive) conformal prediction the practical workhorse \citep{papadopoulos2002icp}. Time-series dependence breaks exchangeability, and a large family of methods restores approximate validity: EnbPI \citep{xu2021enbpi,xu2023enbpi_journal}, adaptive conformal inference and its aggregated form \citep{gibbs2021adaptive,zaffran2022adaptive}, conformalized quantile regression \citep{romano2019cqr}, sequential predictive conformal inference \citep{xu2023spci}, and conformal prediction beyond exchangeability \citep{barber2023beyond}. None delivers an unconditional finite-sample distribution-free guarantee in the general time-series setting; \citet{oliveira2024splitconformal} and \citet{barber2025splitconformal} explain why split conformal is nonetheless effective under temporal dependence. These learned and adaptive methods are exactly the comparators we benchmark the floor against. Our finding is not that they are wrong, but that they clear the trivial floor only narrowly at short horizons --- which makes the floor a binding baseline they should be required to report.

\paragraph{Scoring and calibration.}
For probabilistic forecasts the accepted paradigm is to maximize sharpness subject to calibration \citep{gneiting2007calibration,gneiting2014probabilistic}, using strictly proper scores such as CRPS \citep{gneiting2007strictly,hersbach2000crps} and the Winkler interval score. We use the Winkler score on the large one-step corpus, CRPS on the multi-step seasonal suite, and empirical coverage throughout, following the calibration-then-sharpness protocol of \citet{gneiting2014probabilistic}.

\section{Methods}
\label{sec:methods}
Let $y_1,\ldots,y_T$ be the observed history, $H$ the forecast horizon, $m$ a seasonal period when available, and $\alpha$ the target miscoverage ($\alpha = 0.05$ throughout). All three methods are instances of a single template: a naive point forecast plus the finite-sample split-conformal quantile of its own absolute residuals.

\paragraph{The split-conformal quantile.}
Given nonconformity scores $s_1,\ldots,s_n$ (absolute residuals), the finite-sample $(1-\alpha)$ conformal quantile is
\[
Q_\alpha \;=\; s_{(k)}, \qquad k = \big\lceil (n+1)(1-\alpha) \big\rceil,
\]
the $k$-th order statistic, taken as $+\infty$ if $k > n$ \citep{vovk2005algorithmic,papadopoulos2002icp,lei2018distribution}; it is the standard split-conformal quantile and predates its use for quantile regression by \citet{romano2019cqr}. Ties among residuals need no special handling, and in the online protocol the pool starts from the training residuals and grows by one each step, so $k \le n$ throughout and there is no warm-up. The prediction interval at every horizon is the symmetric band $\mu_h \pm Q_\alpha$ around the naive point $\mu_h$. \emph{Predictive samples}, used for CRPS, come from the same residual pool: we invert the conformal quantile function by reading the band $\mu_h \pm Q_a$ at a grid of levels $a$ and sampling uniformly across levels, giving a piecewise-linear predictive CDF centred at $\mu_h$. Interval (for coverage/Winkler) and samples (for CRPS) thus derive from one construction.

\paragraph{ConformalNaive (random-walk floor).}
Point forecast $\mu_h = y_T$ (the last observed value). Nonconformity scores are the one-step absolute differences $s_t = |y_t - y_{t-1}|$. In an online one-step protocol the point tracks the most recent observation and the residual pool grows as data arrive; in a batch multi-step protocol the point is held flat across the horizon and the interval does not widen with $h$.

\paragraph{ConformalSeasonalNaive (seasonal floor).}
Point forecast $\mu_h = y_{T+h-m}$ (the value one season earlier); scores $s_t = |y_t - y_{t-m}|$; same split-conformal quantile. When $m \le 1$ this reduces to ConformalNaive.

\paragraph{Protocol dependence (the linchpin).}
The behaviour of these floors is sharply protocol-dependent, and the paper's two regimes exploit this. \emph{One step ahead, online}: the last value is an excellent point forecast and the one-step residual quantile is well calibrated, so the floor is strong. \emph{Multi-step, batch}: the point collapses to a single stale value for the whole horizon and the one-step interval does not widen, so the random-walk floor is weak while the seasonal floor partially recovers. We state every result with its protocol so the two regimes cannot be conflated.

\paragraph{ConformalNaive+ (horizon-adaptive, the unifying method).}
The two floors are complementary: the random-walk floor wins at short horizons, the seasonal floor at long ones. ConformalNaive+ selects between them per horizon using a one-line rule computed from the \emph{training history only} (Algorithm~\ref{alg:cnp}). For each lead $h$ it compares the in-sample $h$-step persistence error to the seasonal-lag error and emits the better floor's point, split-conformal interval, and samples. Because long, stale horizons are routed to the seasonal branch, the flat one-step interval is used only where it is valid, which restores calibration across the whole horizon. The rule is not CSP: CSP \citep{manokhin2026csp} mixes a seasonal empirical pool with seasonal residuals and never switches the point forecast by horizon.

\begin{algorithm}[htbp]
\DontPrintSemicolon
\SetKwInOut{KwIn}{Input}
\SetKwInOut{KwOut}{Output}
\KwIn{history $y_{1:T}$, horizon $H$, seasonal period $m$, level $\alpha$}
\KwOut{per-horizon point $\mu_h$, interval $[\,\mu_h - Q, \mu_h + Q\,]$, samples}
\BlankLine
$e_{\mathrm{seas}} \leftarrow \mathrm{median}_{u>m}\, |y_u - y_{u-m}|$\hfill\textit{(seasonal-lag error; flat in $h$)}\;
\For{$h \leftarrow 1$ \KwTo $H$}{
  $e_{\mathrm{rw}}(h) \leftarrow \mathrm{median}_{t}\, |y_{t+h} - y_t|$\hfill\textit{(persistence error; grows with $h$)}\;
  \uIf{$e_{\mathrm{rw}}(h) \le e_{\mathrm{seas}}$}{branch $\leftarrow$ \textsc{ConformalNaive}}
  \Else{branch $\leftarrow$ \textsc{ConformalSeasonalNaive}}
  emit the chosen branch's point, split-conformal interval, and samples at lead $h$\;
}
\caption{ConformalNaive+ (training-only horizon-adaptive selector).}
\label{alg:cnp}
\end{algorithm}

\paragraph{Conformal validity remark.}
The residual pools play the role of a split-conformal calibration set in the spirit of \citet{papadopoulos2002icp}. Split conformal guarantees finite-sample marginal coverage only under exchangeability, which temporal dependence violates, so we make no finite-sample coverage claim and report coverage as an empirical property the methods are judged on. The near-nominal coverage we nonetheless observe (Section~\ref{sec:regime2}) is consistent with recent theory on split conformal beyond exchangeability. Oliveira et al.~\citep{oliveira2024splitconformal} show that split conformal stays valid for non-exchangeable data up to an additive coverage penalty governed by the dependence between the calibration residuals and the test residual; Barber and Pananjady~\citep{barber2025splitconformal} sharpen this for time series, bounding the coverage loss by a mixing-type ``switch'' coefficient that is small for the near-stationary, strongly-mixing series in our suite. The penalty is smallest precisely in the one-step-ahead regime, where the calibration residuals (one-step differences) and the realized one-step error are drawn from nearly the same distribution; it grows with horizon as the flat interval is reused against an increasingly stale point forecast --- which is exactly the coverage decay we measure in Figure~\ref{fig:horizon} and the failure ConformalNaive+ routes around. This characterization also clarifies the contrast with DeepNPTS: the floor's intervals are anchored to a calibration set whose validity degrades gracefully and quantifiably under dependence, whereas the neural model's intervals carry no such anchor and undercover by a third. Foregoing a hard guarantee is the norm in the time-series conformal literature \citep{xu2021enbpi,gibbs2021adaptive,barber2023beyond}.

\section{Experimental Design}
\label{sec:design}
We use two benchmarks with two protocols, deliberately chosen to expose both regimes of the floor.

\paragraph{Regime~1 --- one-step-ahead online ($2{,}217$ series).}
We use $2{,}217$ real series drawn from nine public sources --- the Monash forecasting archive \citep{godahewa2021monash}, the LOTSA collection \citep{woo2024moirai}, the LTSF traffic/electricity/weather benchmark suites, METR-LA, BOOM, and nips/probts --- spanning minutely to daily frequencies (Appendix~\ref{app:manifest}, Table~\ref{tab:manifest}, lists every source, its retained series count, and seasonal periods). Of $2{,}373$ candidate series, $2{,}217$ are retained; $156$ are dropped for being shorter than $T_{\mathrm{train}}+T_{\mathrm{test}}=1{,}100$ observations plus lag context. The seasonal period $m$ for each series is the integer value recorded in the source metadata, not estimated; series are used as provided, with no imputation, normalisation, or frequency conversion. Each series is forecast one step at a time online over a $300$-step test span ($T_{\mathrm{train}}=800$), revealing the actual after each step and updating the residual pool. The metric is the Winkler interval score (lower is better). We compare the two floors against $19$ methods spanning classical naive-interval baselines (NaiveInterval, NaiveDiffInterval, SeasonalNaiveInterval), the NPTS family (NPTS with mild and strong recency-decay kernels, and SeasonalNPTS), the published CSP method, and a battery of learned conformal predictors (CQR, RCI, EnbPI, EnsCQR, KOWCPI, QR, QEns, NexCP, ACI, AgACI, SPCI, WeightedConformal). Every comparator is run in this same online one-step pipeline on the same series at its reference-implementation defaults with no per-dataset tuning, each applying its own online update rule from its reference implementation (the learned conformal predictors share a ridge base learner on lag features), so every method is evaluated identically rather than re-tuned. The headline statistic is the per-series win rate, which compares each method against ConformalNaive on the same series and is invariant to per-series scale; as a scale-free effect size we report the median per-series relative Winkler difference $(W_{\mathrm{floor}}-W_{\mathrm{comp}})/W_{\mathrm{comp}}$. Significance uses a paired Wilcoxon signed-rank test, but at $n=2{,}217$ it is uninformative for ranking (almost every comparison is significant regardless of effect size), so we band comparators by win rate and effect size and apply Holm--Bonferroni only to the headline claims, which it leaves unchanged.

\paragraph{Regime~2 --- multi-step seasonal (6 datasets).}
The six GluonTS datasets on which DeepNPTS was introduced \citep{rangapuram2023deepnpts} (electricity, exchange\_rate, solar\_energy, taxi, traffic, wikipedia), under the audited rolling-origin protocol of the companion CSP study \citep{manokhin2026csp}: $H = 24$ (hourly) or $30$ (daily), $B = 100$ samples, $\alpha = 0.05$, seed $0$, $380$ forecast records per method. The metric is empirical CRPS with empirical $95\%$ coverage. Our three methods are scored on the identical windows and paired per-(dataset, series, window) against the audited comparators --- NPTS, SeasonalNPTS, CSP-Adaptive, CSP-Fixed, and a protocol-matched DeepNPTS rerun --- so per-window paired Wilcoxon tests apply directly.

\section{Results}
\label{sec:results}

\subsection{The headline: one step ahead}
\label{sec:regime1}
Table~\ref{tab:regime1} sorts the $19$ comparators into three bands by ConformalNaive's per-series win rate over the $2{,}217$ series; Figure~\ref{fig:regime1} shows the full ranking. ConformalNaive \emph{decisively beats} the naive value-quantile baselines (NaiveInterval $90\%$, SeasonalNaiveInterval $92\%$), the entire NPTS family (NPTS\_strong $73\%$, SeasonalNPTS $64\%$), and the published CSP method ($71\%$, bootstrap $95\%$ CI $[69,73]$, paired Wilcoxon $p \approx 7.6 \times 10^{-135}$). This is not an artifact of corpus composition: ConformalNaive beats CSP on $7$ of the $9$ sources and NPTS\_strong on $7$ of $9$ (Appendix~\ref{app:bysource}, Table~\ref{tab:bysource}). It is \emph{statistically on par} with the simpler learned conformal predictors (RCI and quantile regression: win rate $47$--$48\%$, median relative Winkler within $2\%$). It \emph{trails} the adaptive-online and ensemble methods --- precisely those that adapt the quantile to distribution shift --- by a real but bounded margin: SPCI, EnsCQR, AgACI, ACI, and CQR win $56$--$76\%$ of series and lead by a median $9$--$33\%$ relative Winkler, with wide spread (e.g.\ SPCI median $+29\%$, inter-quartile range $[+1\%, +68\%]$; Table~\ref{tab:regime1full}). The pattern is interpretable: last-value-plus-conformal matches static split-conformal predictors and is beaten specifically by shift-tracking methods, not by the bulk of the conformal toolkit. This is the binding-baseline result: the naive value-quantile baselines, the entire NPTS family, and the published CSP method are all beaten by the floor, so a learned method that does not at least clear it has not earned its complexity. (At $n=2{,}217$ the paired Wilcoxon $p$ is below any threshold for nearly every comparator --- a function of sample size, not effect size --- so we band comparators by win rate and median relative Winkler, not by significance; Holm--Bonferroni correction across the comparisons leaves the headline results unchanged.) Crucially, the floor's one-step strength is not bought by undercoverage: ConformalNaive's mean empirical coverage over the $2{,}217$ series is $0.91$ (nominal $0.95$), at or above every comparator --- including the adaptive methods that beat it on Winkler (SPCI $0.89$, ACI $0.90$, CQR $0.86$) and CSP ($0.85$) --- so it is simultaneously well-calibrated and competitive, not sharp-but-miscalibrated. Nor are its intervals artificially narrow: against the adaptive methods that lead on Winkler it is in fact $13$--$36\%$ \emph{wider} (median relative width) yet better covered, and against CSP and NPTS it is both narrower and better calibrated.

\begin{table}[htbp]
\floatconts
  {tab:regime1}
  {\caption{One-step-ahead head-to-head: ConformalNaive vs.\ $19$ comparators over $2{,}217$ real series (Winkler score). Each comparator is annotated with ConformalNaive's per-series win rate; bands are by win rate. All ``Beats'' entries have paired Wilcoxon $p < 10^{-86}$.}}
  {\resizebox{\linewidth}{!}{\begin{tabular}{p{2.2cm}p{7.4cm}rr}
\toprule
Band & Comparators (win \% in subscript) & win \% & med.\ rel.\ $\Delta W$ \\
\midrule
\textbf{Beats} & SeasonalNaiveInterval~92, NaiveInterval~90, NPTS\_mild~86, NPTS\_strong~73, CSP~71, SeasonalNPTS~64 & 64--92 & -42 to -20\% \\
\addlinespace
\textbf{On par} & NaiveDiffInterval~49, QR~48, RCI~47 & 47--49 & +0 to +2\% \\
\addlinespace
\textbf{Trails} & CQR~44, KOWCPI~41, NexCP~39, EnbPI~39, ACI~33, QEns~33, WeightedConformal~30, EnsCQR~28, AgACI~26, SPCI~24 & 24--44 & +5 to +33\% \\
\bottomrule
\end{tabular}
}}
\end{table}

\begin{figure}[htbp]
\floatconts
  {fig:regime1}
  {\caption{ConformalNaive win rate against every comparator, one step ahead, $2{,}217$ series. Green: ConformalNaive beats the comparator on a majority of series; grey: statistical par ($45$--$55\%$); red: ConformalNaive trails. The dashed line is the $50\%$ tie. The published CSP method and the entire NPTS family fall in the green band.}}
  {\includegraphics[width=0.86\linewidth]{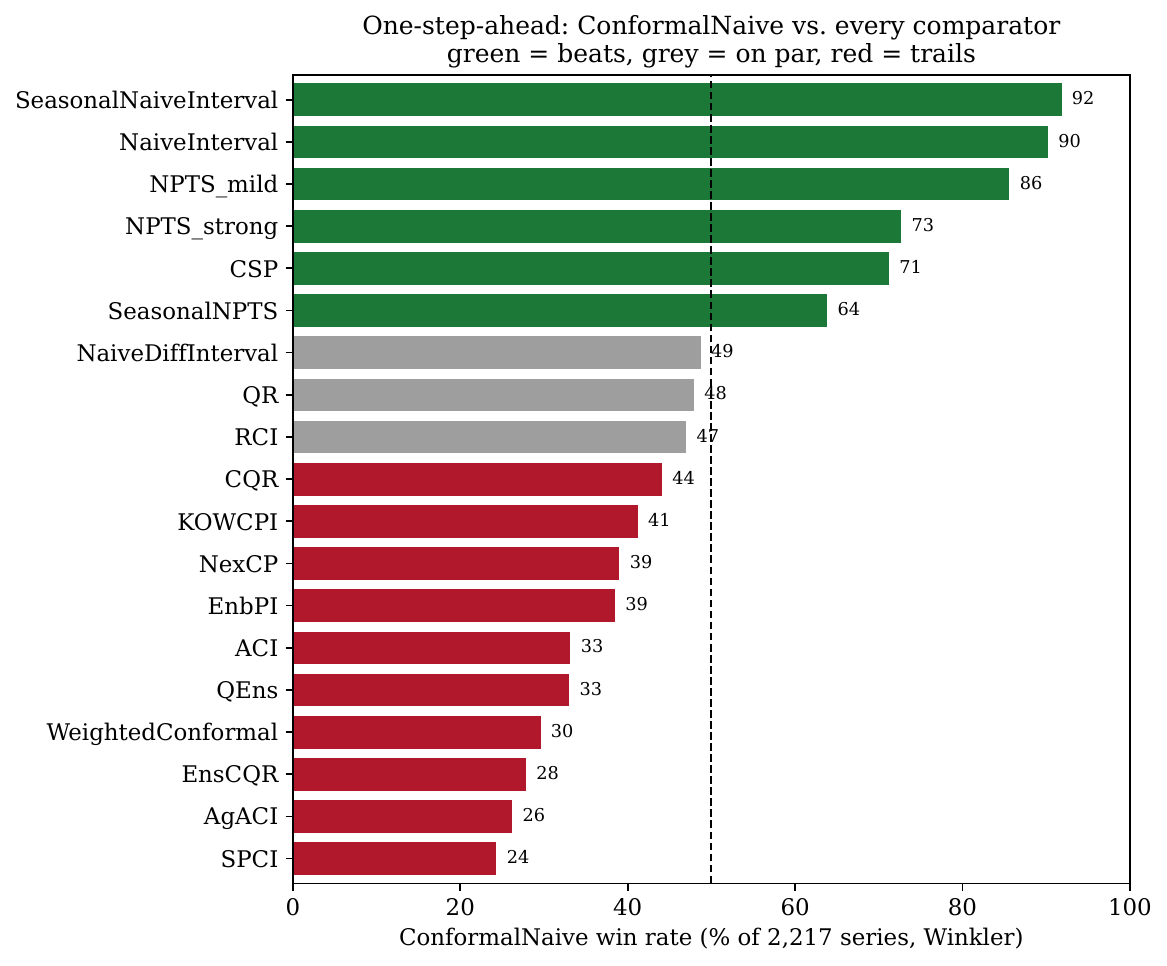}}
\end{figure}

The seasonal floor behaves oppositely at one step: ConformalSeasonalNaive edges only the three classical baselines ($52$--$55\%$) and loses to everything else (CSP $29\%$, learned methods $2$--$12\%$), because the season-ago value is a poor one-step point forecast and its seasonal residuals give very wide intervals (full table in Appendix~\ref{app:regime1full}). The two floors are mirror images, a fact we exploit in Section~\ref{sec:crossover}.

\subsection{The boundary: multi-step seasonal}
\label{sec:regime2}
Table~\ref{tab:regime2} reports the multi-step seasonal regime, where the floor is \emph{not} the answer and we say so plainly. CSP clearly wins on CRPS rank; ConformalNaive+ and ConformalSeasonalNaive beat NPTS and tie the protocol-matched DeepNPTS rerun, but trail CSP; the random-walk floor is last. Complexity earns its keep exactly here --- this is the boundary of the headline claim. The aggregate ranks hide dataset-level variation (e.g.\ SeasonalNPTS is best on electricity but near-worst on exchange\_rate); the per-dataset breakdown is in Appendix~\ref{app:bydataset}, Table~\ref{tab:bydataset}.

\begin{table}[htbp]
\floatconts
  {tab:regime2}
  {\caption{Multi-step seasonal regime (6 datasets, $380$ windows, CRPS). Lower rank is better; coverage targets $0.95$. Our methods in bold. CSP wins; the random-walk floor is last; DeepNPTS --- the only trained method --- is the worst calibrated.}}
  {\begin{tabular}{lrr}
\toprule
Method & CRPS rank (of 8) & Coverage \\
\midrule
CSP-Adaptive & 3.49 & 0.89 \\
CSP-Fixed & 3.59 & 0.89 \\
SeasonalNPTS & 4.00 & 0.91 \\
DeepNPTS & 4.46 & 0.66 \\
\textbf{ConformalNaive+} & 4.61 & 0.84 \\
\textbf{ConformalSeasonalNaive} & 4.74 & 0.85 \\
NPTS & 5.37 & 0.95 \\
\textbf{ConformalNaive} & 5.74 & 0.76 \\
\bottomrule
\end{tabular}
}
\end{table}

\begin{figure}[htbp]
\floatconts
  {fig:regime2}
  {\caption{Mean CRPS rank, multi-step seasonal regime (lower is better). Our three methods in orange. CSP-Adaptive leads; ConformalNaive+ is the best of the three floors; the random-walk floor is last.}}
  {\includegraphics[width=0.78\linewidth]{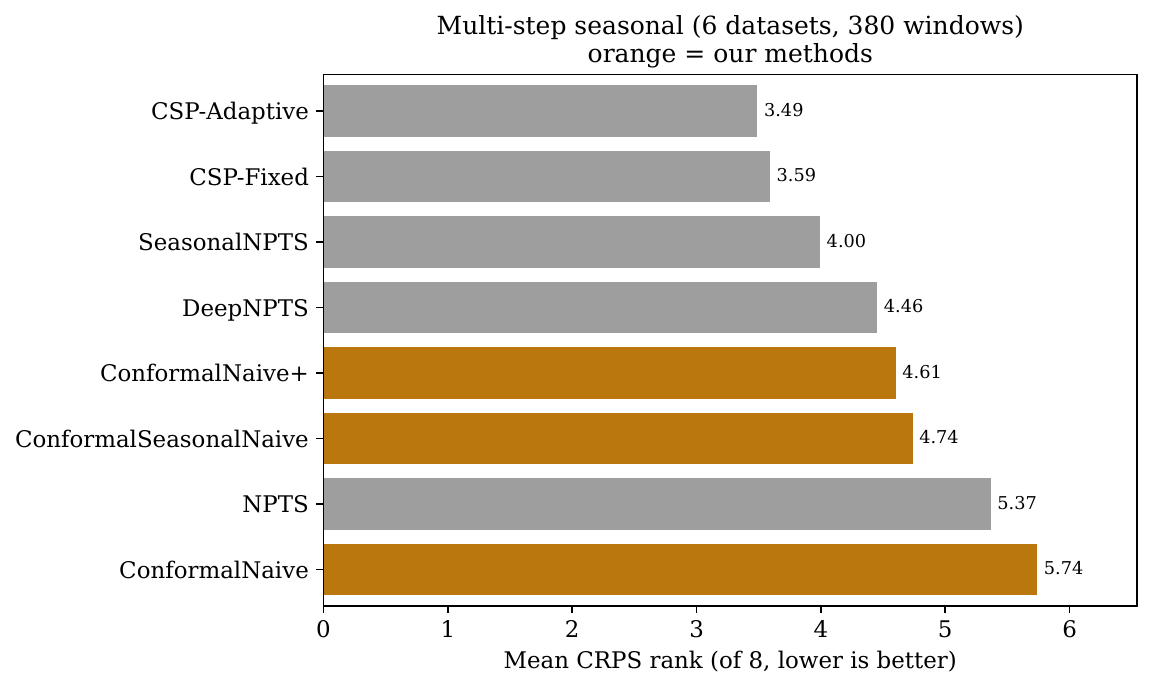}}
\end{figure}

\paragraph{Calibration: the finding that survives.}
The accuracy ranking is the floor's limit; the calibration ranking is a separate result that holds regardless. Figure~\ref{fig:calibration} shows that DeepNPTS, the only trained method in the comparison, is the \emph{worst}-calibrated: its nominal-$95\%$ interval covers the truth only $66\%$ of the time, against $84$--$85\%$ for the trivial conformal floors and $0.89$--$0.95$ for the NPTS family. Under the calibration-then-sharpness paradigm of \citet{gneiting2014probabilistic}, an interval that misses a third of the time fails the basic prerequisite. A one-line conformal interval is more honest about its uncertainty than the neural baseline, on the neural baseline's own evaluation suite. (NPTS reaches $0.95$ coverage almost by construction --- it samples the whole empirical history, giving wide, near-unconditional intervals that are well calibrated but not sharp, which is why it is near-last on CRPS rank.)

\begin{figure}[htbp]
\floatconts
  {fig:calibration}
  {\caption{Empirical $95\%$ coverage, multi-step seasonal regime (target $0.95$, dashed). DeepNPTS (red), the only trained method, is the worst calibrated at $0.66$; the trivial conformal floors (orange) sit at $0.84$--$0.85$.}}
  {\includegraphics[width=0.78\linewidth]{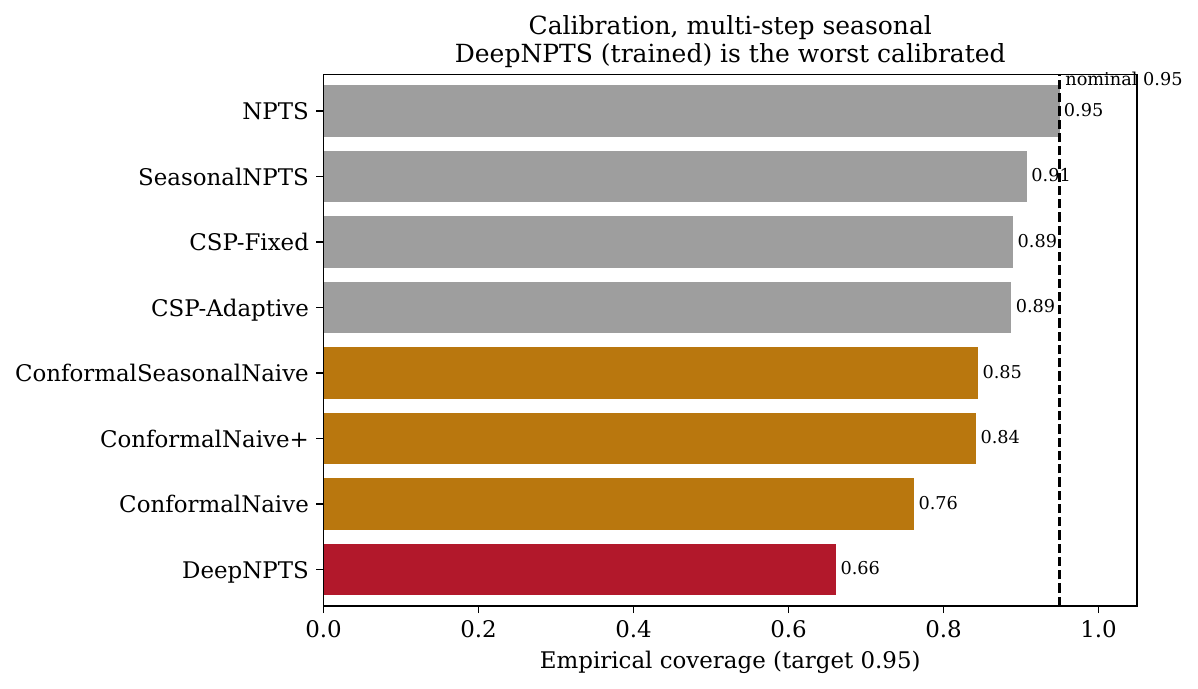}}
\end{figure}

Table~\ref{tab:regime2wil} gives the per-window paired Wilcoxon tests of our three methods against the key comparators. ConformalNaive+ significantly beats NPTS ($64\%$ of windows, $p \approx 0$) and edges the matched DeepNPTS rerun ($51\%$, $p = 7 \times 10^{-3}$); ConformalSeasonalNaive beats NPTS and ties DeepNPTS; both lose to CSP; the random-walk floor loses to all.

\begin{table}[htbp]
\floatconts
  {tab:regime2wil}
  {\caption{Per-window paired Wilcoxon, multi-step seasonal. Each cell: \% of windows our method beats the comparator on CRPS / one-sided $p$ that our method is better (n.s.\ $= p>0.05$).}}
  {\resizebox{\linewidth}{!}{\begin{tabular}{lrrr}
\toprule
Our method & NPTS & DeepNPTS & CSP-Adaptive \\
\midrule
ConformalNaive & 33\%/n.s. & 32\%/n.s. & 24\%/n.s. \\
ConformalSeasonalNaive & 62\%/6e-14 & 49\%/n.s. & 29\%/n.s. \\
ConformalNaive+ & 64\%/$\approx$0 & 51\%/7e-3 & 32\%/n.s. \\
\bottomrule
\end{tabular}
}}
\end{table}

\subsection{The horizon crossover and ConformalNaive+}
\label{sec:crossover}
Why is the random-walk floor first at one step and last at $24$? Figure~\ref{fig:horizon} sweeps the horizon on the four hourly datasets. The two floors cross over at $h \approx 2$--$3$: ConformalNaive is best at $h=1$ but its error balloons mid-cycle (the phase where the last value is most stale) and its coverage collapses from $0.99$ to $0.64$, while ConformalSeasonalNaive holds near-nominal coverage throughout. ConformalNaive+ (Algorithm~\ref{alg:cnp}) tracks the lower of the two at every horizon and --- because it routes stale horizons to the seasonal branch --- restores coverage to $0.85$--$0.99$ across the whole horizon. It does not reach the per-window oracle envelope (best-of-two), since it selects per horizon rather than per window. Closing that gap is future work: a per-series rule (applying the same in-sample error test on each series' own history), an online per-window selector that tracks each floor's recent interval score, or a soft blend weighted by recent error would all approach the oracle while staying training-free --- at the cost of the one-line simplicity that motivates ConformalNaive+.

\begin{figure}[htbp]
\floatconts
  {fig:horizon}
  {\caption{Horizon sweep (four hourly datasets, $H=24$). \emph{Top:} CRPS normalized to ConformalNaive at $h=1$. The floors cross over at $h\approx 2$--$3$; ConformalNaive+ (gold) tracks the better floor; the dotted line is the per-window oracle envelope. \emph{Bottom:} coverage. ConformalNaive (red) collapses mid-horizon; ConformalNaive+ and the seasonal floor stay near nominal.}}
  {\includegraphics[width=0.9\linewidth]{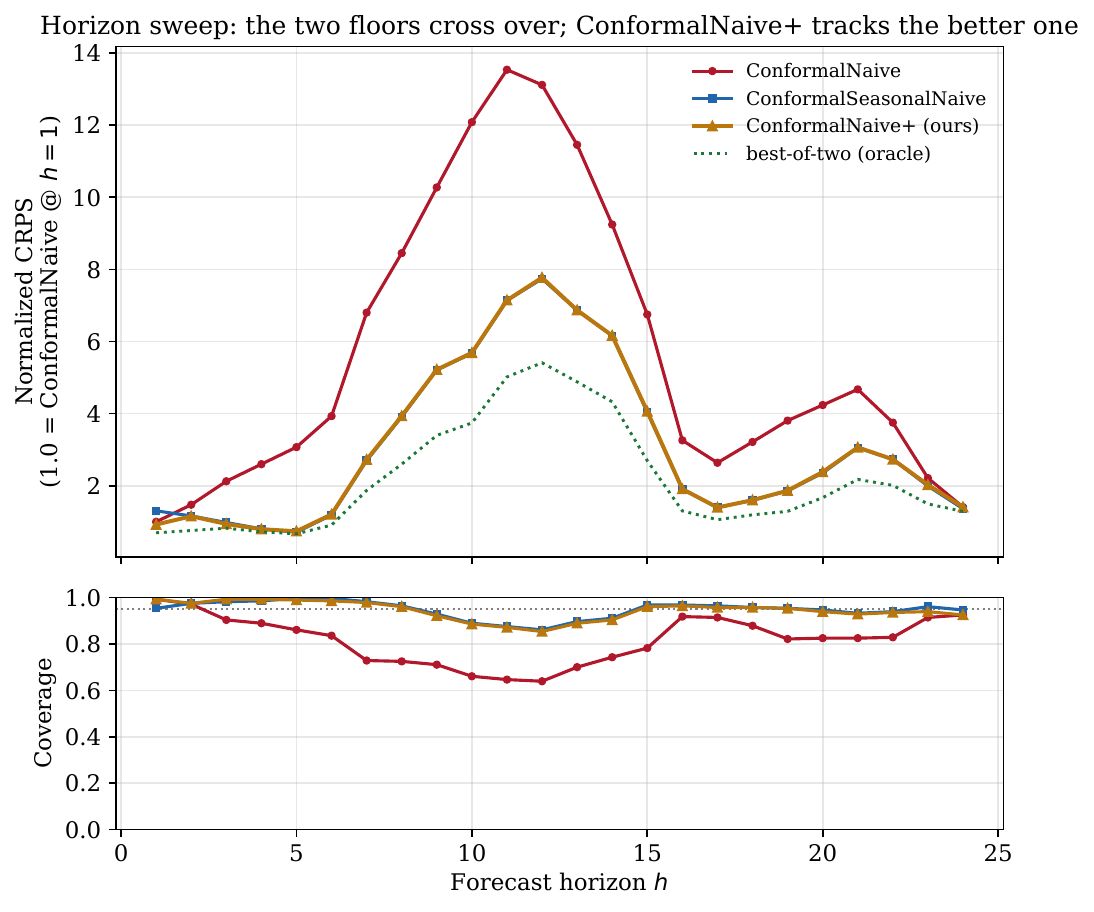}}
\end{figure}

\section{What Works, What Does Not, and Where}
\label{sec:diagnosis}
The results assemble into a compact regime map. At short horizons the random-walk floor is near the top of the field --- beating the NPTS family and CSP and beaten only by the adaptive-online methods (by a bounded relative margin) --- and the seasonal floor is weak. At multi-step seasonal horizons the picture inverts: the random-walk floor is last, the seasonal floor is mid-pack, and the seasonal pool (CSP) wins. The two floors are mirror images, and ConformalNaive+ unifies them with a one-line rule. The practitioner decision rule follows directly: at horizon $\approx 1$, report and use ConformalNaive; at multi-step seasonal horizons, report ConformalSeasonalNaive as the floor but expect CSP to win; if unsure, use ConformalNaive+. In every case, the matching floor is the baseline a learned method must clear before its gains are credible. The calibration result cuts across the map and is independent of the accuracy ranking: DeepNPTS --- the only trained method we test --- is the worst-calibrated of all eight methods on the multi-step seasonal suite (its own home regime), covering the truth only $66\%$ of the time at a nominal $95\%$ against $84$--$85\%$ for the trivial floors.

\section{Limitations and Scope}
\label{sec:limits}
\paragraph{Not a SOTA claim.} We do not claim a new best forecaster. In the multi-step seasonal regime the floors lose to CSP and only match DeepNPTS; that boundary is reported, not hidden. The win over CSP at one step ($71\%$) is CSP evaluated outside its multi-step seasonal design regime --- CSP is top of its own six-dataset suite (Table~\ref{tab:regime2}), and the one-step result is not evidence against CSP but against omitting the floor.
\paragraph{Two corpora.} The regime map is read across two benchmarks (the $2{,}217$-series one-step corpus and the six-dataset multi-step suite) rather than one factorial design; the horizon sweep (Section~\ref{sec:crossover}) partly closes this by varying the horizon within one suite.
\paragraph{Construction.} The floors treat horizons independently within a predictive sample, so columns are not coherent trajectories; the one-step interval does not widen with $h$ (which is precisely the failure ConformalNaive+ routes around); and residual exchangeability holds only approximately under dependence, so coverage is empirical, not guaranteed.
\paragraph{Comparator scope and protocol.} All comparators run at default settings in one shared pipeline; we neither tuned them per dataset nor reimplemented them. The result should therefore be read as ``these methods, at standard settings, do not clear the floor at one step,'' not as a statement about their best-tuned performance. The DeepNPTS calibration result is for the audited configuration of the companion CSP study (Appendix~\ref{app:impl}) and characterises that configuration, not a fully re-tuned DeepNPTS.
\paragraph{Application scope.} The study is univariate and covers short-horizon online and multi-step seasonal forecasting. It does not address covariates, hierarchical or intermittent demand, multivariate trajectories, cold-start series, or long-horizon non-seasonal forecasting, where the floor's standing is untested; the title's ``mandatory baseline'' claim is intended for the probabilistic time-series settings studied here.
\paragraph{Scores and robustness, left to future work.} The floors use a symmetric absolute-residual score; asymmetric or scale-robust (MAD-studentised) variants may fit skewed or heteroskedastic series better. We did not sweep $\alpha$, train/test lengths, or updating vs.\ rolling residual pools, and we did not include classical distributional baselines (ETS/ARIMA with bootstrapped residuals) or additional neural comparators such as Temporal Fusion Transformers in the multi-step regime; these are natural extensions. ConformalNaive+ uses the median in-sample error for outlier-robustness; mean-based, rolling, or per-series selection variants are untested and could narrow the gap to the per-window oracle.

\section{Conclusion}
\label{sec:conclusion}
A last-value point forecast wrapped in a finite-sample split-conformal residual quantile --- no parameters, no training --- beats the published CSP method and the entire NPTS family on $2{,}217$ real series at one step ahead, matches the simpler learned conformal predictors and is beaten only by shift-tracking methods, and is better calibrated than a trained neural forecaster on that forecaster's own datasets. It is not state of the art: at multi-step seasonal horizons it is the weakest method, and the seasonal pool (CSP) wins. The contribution is therefore one of benchmarking discipline rather than a leaderboard: this floor, and its seasonal and horizon-adaptive variants, must be reported whenever a learned probabilistic forecaster claims gains, and probabilistic-forecasting benchmarks should report empirical coverage at the nominal level alongside any sharpness score. The methods that clear the floor are specifically the adaptive, shift-tracking conformal predictors (leading by $9$--$33\%$ relative Winkler), and even they do so only at the short horizons where the floor is designed to be strong.

\acks{The author thanks the maintainers of the GluonTS, Monash, and LOTSA resources for the public datasets used in this benchmark.}

\paragraph{Data and code availability.} The manuscript source, the analysis scripts, and the full per-series and per-window result tables underlying every figure and table in this paper are archived at Zenodo: \url{https://doi.org/10.5281/zenodo.20594485}.

\appendix

\section{Method and Protocol Details}
\label{app:impl}
\paragraph{Floors.} ConformalNaive: point $y_T$, scores $|y_t - y_{t-1}|$, split-conformal quantile at $\alpha=0.05$. ConformalSeasonalNaive: point $y_{T+h-m}$, scores $|y_t - y_{t-m}|$. ConformalNaive+: Algorithm~\ref{alg:cnp}; the per-horizon selection uses median in-sample $h$-step persistence error vs.\ median seasonal-lag error.
\paragraph{Regime~1 protocol.} One-step-ahead online with batched residual updates, $T_{\mathrm{train}}=800$, $T_{\mathrm{test}}=300$; Winkler score (lower is better); the per-series win rate is invariant to scale, so no normalization is applied; $2{,}217$ series across nine sources (full manifest in Appendix~\ref{app:manifest}, Table~\ref{tab:manifest}).
\paragraph{Regime~2 protocol.} Rolling-origin, $H \in \{24,30\}$, $B=100$, $\alpha=0.05$, seed $0$, $380$ records/method; CRPS and empirical coverage; paired against the audited comparators of the companion CSP study, including a protocol-matched DeepNPTS rerun ($50$ epochs, $200$ inference samples, CPU).

\section{Dataset Manifest (One-Step Corpus)}
\label{app:manifest}
Table~\ref{tab:manifest} lists the $2{,}217$ retained series of the one-step corpus by source, with the seasonal periods present in each. All series come from public archives; seasonal periods are taken from dataset metadata. Series shorter than $T_{\mathrm{train}}+T_{\mathrm{test}}=1{,}100$ observations (plus lag context) are excluded ($156$ of $2{,}373$ candidates).

\begin{table}[htbp]
\floatconts
  {tab:manifest}
  {\caption{One-step corpus by source: retained series and the seasonal periods $m$ present. Total $2{,}217$ series.}}
  {\begin{tabular}{lrl}
\toprule
Source & Series & Seasonal periods $m$ \\
\midrule
traffic-LTSF      & 862 & 24 \\
LOTSA             & 440 & 7, 52, 96, 144, 288 \\
electricity-LTSF  & 321 & 24 \\
Monash archive    & 272 & 1, 7, 12, 24, 48, 52, 144 \\
METR-LA           & 207 & 288 \\
BOOM              & 50  & 24, 48, 288, 1440, 8640 \\
tier-A (mixed)    & 36  & 7, 24, 96 \\
weather-LTSF      & 21  & 144 \\
nips/probts       & 8   & 7 \\
\midrule
\textbf{Total}    & \textbf{2,217} & \\
\bottomrule
\end{tabular}
}
\end{table}

\section{Full One-Step Head-to-Head}
\label{app:regime1full}
Table~\ref{tab:regime1full} reports both floors against all $19$ comparators on the $2{,}217$-series one-step corpus: per-series win rate and one-sided paired Wilcoxon $p$ (that the floor is better).

\begin{table}[htbp]
\floatconts
  {tab:regime1full}
  {\caption{Full one-step head-to-head over $2{,}217$ series. CN $=$ ConformalNaive, CSN $=$ ConformalSeasonalNaive. Win \% is the share of series on which the floor has the lower Winkler score; $p$ is the one-sided paired Wilcoxon $p$-value that the floor is better.}}
  {\resizebox{0.8\linewidth}{!}{\begin{tabular}{lrrrr}
\toprule
Comparator & CN win \% & CN med.\ rel.\ $\Delta W$ (\%) & CSN win \% & CSN med.\ rel.\ $\Delta W$ (\%) \\
\midrule
SeasonalNaiveInterval & 91.9 & -36.0 & 54.6 & -0.1 \\
NaiveInterval & 90.3 & -42.2 & 52.3 & -2.7 \\
NPTS\_mild & 85.7 & -32.7 & 39.1 & +12.9 \\
NPTS\_strong & 72.7 & -21.7 & 25.7 & +38.2 \\
CSP & 71.3 & -25.9 & 29.1 & +21.8 \\
SeasonalNPTS & 63.9 & -20.4 & 27.9 & +27.8 \\
NaiveDiffInterval & 48.8 & +0.0 & 7.2 & +56.3 \\
QR & 48.0 & +2.4 & 8.8 & +71.7 \\
RCI & 47.0 & +1.7 & 12.5 & +58.6 \\
CQR & 44.1 & +9.4 & 6.1 & +78.4 \\
KOWCPI & 41.2 & +5.3 & 3.2 & +62.7 \\
NexCP & 39.1 & +6.4 & 6.9 & +69.0 \\
EnbPI & 38.5 & +6.1 & 3.6 & +60.9 \\
ACI & 33.2 & +9.5 & 3.4 & +69.3 \\
QEns & 33.0 & +15.4 & 3.0 & +78.1 \\
WeightedConformal & 29.6 & +12.0 & 6.0 & +77.3 \\
EnsCQR & 27.9 & +32.9 & 6.0 & +105.4 \\
AgACI & 26.3 & +13.1 & 2.8 & +77.6 \\
SPCI & 24.4 & +29.0 & 2.1 & +117.6 \\
\bottomrule
\end{tabular}
}}
\end{table}

\section{Per-Source Robustness (One-Step)}
\label{app:bysource}
Table~\ref{tab:bysource} reports ConformalNaive's win rate against the key comparators within each of the nine sources, showing the headline is not driven by any single corpus: ConformalNaive beats CSP on $7/9$ sources and NPTS on $7/9$, and is beaten by SPCI on $7/9$ (consistent with the aggregate).

\begin{table}[htbp]
\floatconts
  {tab:bysource}
  {\caption{One-step win rate (\%) of ConformalNaive by source against key comparators. Bottom row is the pooled corpus.}}
  {\begin{tabular}{lrrrrr}
\toprule
Source & Series & vs CSP & vs NPTS\_strong & vs SeasonalNPTS & vs SPCI \\
\midrule
traffic-LTSF & 862 & 58 & 81 & 37 & 4 \\
LOTSA & 440 & 87 & 63 & 82 & 33 \\
electricity-LTSF & 321 & 81 & 94 & 85 & 43 \\
Monash & 272 & 65 & 68 & 68 & 16 \\
METR-LA & 207 & 87 & 41 & 93 & 71 \\
BOOM & 50 & 46 & 20 & 52 & 16 \\
tier-A & 36 & 83 & 86 & 78 & 56 \\
weather-LTSF & 21 & 95 & 86 & 95 & 0 \\
nips/probts & 8 & 100 & 100 & 100 & 100 \\
\midrule
\textbf{All} & \textbf{2217} & \textbf{71} & \textbf{73} & \textbf{64} & \textbf{24} \\
\bottomrule
\end{tabular}
}
\end{table}

\section{Per-Dataset Multi-Step Breakdown}
\label{app:bydataset}
Table~\ref{tab:bydataset} gives per-dataset mean CRPS rank (of eight methods) in the multi-step seasonal regime, behind the aggregate of Table~\ref{tab:regime2}.

\begin{table}[htbp]
\floatconts
  {tab:bydataset}
  {\caption{Per-dataset mean CRPS rank (of 8) on the six-dataset multi-step suite. Lower is better; our methods in bold.}}
  {\resizebox{\linewidth}{!}{\begin{tabular}{lrrrrrrr}
\toprule
Method & elect & excha & solar & taxi & traff & wikip & Mean \\
\midrule
CSP-Adaptive & 3.7 & 2.7 & 3.6 & 4.4 & 2.8 & 3.8 & 3.49 \\
SeasonalNPTS & 2.5 & 5.6 & 3.9 & 4.0 & 3.2 & 4.8 & 4.00 \\
DeepNPTS & 4.7 & 5.4 & 4.1 & 3.8 & 5.7 & 3.0 & 4.46 \\
\textbf{ConformalNaive+} & 4.1 & 4.6 & 4.8 & 5.1 & 4.0 & 5.1 & 4.61 \\
\textbf{ConformalSeasonalNaive} & 4.3 & 4.6 & 4.4 & 5.9 & 4.0 & 5.2 & 4.74 \\
\textbf{ConformalNaive} & 6.9 & 4.3 & 5.9 & 4.9 & 7.2 & 5.2 & 5.74 \\
\bottomrule
\end{tabular}
}}
\end{table}

\bibliographystyle{plainnat}
\bibliography{refs}

\end{document}